\documentclass[conference]{IEEEtran}
\usepackage[latin9]{inputenc}
\usepackage{units}
\usepackage{amsmath,amsthm}
\usepackage{amstext}
\usepackage{amssymb,amsfonts}
\usepackage{graphicx}
\usepackage[caption=false]{subfig}
\usepackage{epstopdf}

\makeatletter
\theoremstyle{plain}
\newtheorem{thm}{\protect\theoremname}
\newtheorem{assumption}[thm]{Assumption}

\newcommand{\D}{\mathbf{D}}

\newcommand{\x}{\mathbf{x}}
\newcommand{\C}{\mathbf{C}}
\renewcommand{\c}{\mathbf{c}} 

\newcommand{\y}{\mathbf{y}}

\newcommand{\R}{\mathbf{R}} 
\newcommand{\RR}{\mathbb{R}} 
\newcommand{\I}{\mathcal{I}}

\ifCLASSINFOpdf
\else
\fi
\usepackage[noadjust]{cite}
\author{\IEEEauthorblockN{Farhad Pourkamali-Anaraki\IEEEauthorrefmark{1}\IEEEauthorrefmark{3},
Stephen Becker\IEEEauthorrefmark{2},
Shannon M. Hughes\IEEEauthorrefmark{1}}
\IEEEauthorblockA{\IEEEauthorrefmark{1}Department of Electrical, Computer, and Energy Engineering\\
University of Colorado at Boulder,
Boulder, Colorado 80309}
\IEEEauthorblockA{\IEEEauthorrefmark{2}Department of Applied Mathematics\\
University of Colorado at Boulder,
Boulder, Colorado 80309}
\IEEEauthorblockA{\IEEEauthorrefmark{3}Email: Farhad.Pourkamali@colorado.edu}
}

\makeatother

\providecommand{\theoremname}{Theorem}

\begin{document}

\title{Efficient Dictionary Learning via Very Sparse Random Projections}
\maketitle
\begin{abstract}
Performing signal processing tasks on compressive measurements of data has received great attention in recent years. In this paper, we extend previous work on compressive dictionary learning by showing that more general random projections may be used, including sparse ones.  More precisely, we examine compressive K-means clustering as a special case of compressive dictionary learning and give theoretical guarantees for its performance for a very general class of random projections. We then propose a memory and computation efficient dictionary learning algorithm, specifically designed for analyzing large volumes of high-dimensional data, which learns the dictionary from very sparse random projections. Experimental results demonstrate that our approach allows for reduction of computational complexity and memory/data access, with controllable loss in accuracy.
\end{abstract}
\vspace{-0.1in}
\section{Introduction}

There are several ways to represent low-dimensional structure of high-dimensional
data, the best known being principal component analysis (PCA).
However, PCA is based on a linear subspace model that is generally
 not capable of capturing the geometric structure of real-world datasets
\cite{WakinLowDim}.

The sparse signal model is a nonlinear generalization of the linear
subspace model that has been
used in various signal and image processing tasks \cite{OnTheRoleSparse,TaskDrivenDictLearn,NIPS_Marial},
as well as compressive sensing \cite{CandesIntroCS}.
This model assumes
 that each data sample can be represented as a linear
combination of a few elements (atoms) from a dictionary. 
Data-adaptive dictionary learning can lead 
 to a much more compact representation than predefined dictionaries
such as wavelets, and 
thus a central problem is finding a good data-adaptive dictionary.

Dictionary learning algorithms such as the method of optimal directions
(MOD) \cite{MOD} and the K-SVD algorithm \cite{KSVD} aim
to learn a dictionary by minimizing the representation error of data
in an iterative procedure involving two steps of sparse coding
and dictionary update. The latter often requires ready
access to the entire data available at a central processing unit. 

Due to increasing sizes of datasets, not only do algorithms take longer to run,
but 
 it may not even be feasible
or practical to acquire and/or hold every data entry.
In applications such as distributed databases, where data is typically
distributed over an interconnected set of distributed sites \cite{CloudKSVD},
it is important to avoid communicating the entire data.

A promising approach to address these issues is to take a compressive sensing approach, where we only have access to compressive measurements of data. In fact, performing signal processing and data mining tasks
on compressive versions of the data has been an important topic in
the recent literature. For example, in \cite{SignalProcCompMeas},
certain inference problems such as detection and estimation within the compressed domain have been studied. Several lines of work consider recovery of principal components \cite{Fowler,InvariancePCsHanchao,CPCA_ICASSP_Farhad,Farhad_ICML}, spectral features \cite{SketchedSVD}, and change detection \cite{ChangeDetection} from compressive measurements.
 
In this paper, we focus on the problem of dictionary learning based
on compressive measurements. Our contributions
are twofold. 
First, we show the connection between dictionary learning in the compressed domain and K-means clustering.
Most standard dictionary learning
algorithms are indeed a generalization of the K-means clustering algorithm
\cite{HastieElementsStat}, where the reference to K-means is a common
approach to analyze the performance of these algorithms \cite{KSVD,RLS_DLA}.
 This paper takes initial steps towards providing theoretical guarantees
for recovery of the true underlying dictionary from compressive measurements. Moreover, our analysis applies to compressive measurements
obtained by a general class of random matrices consisting of i.i.d.\ zero-mean
entries and finite first four moments.

Second, we extend the prior work in \cite{CKSVD_ICASSP} where compressive dictionary learning for random Gaussian matrices is considered. In particular, we propose a memory and computation efficient dictionary learning
algorithm applicable to modern data settings. To do this, we learn a dictionary from very sparse random
projections, i.e.~projection of the data onto a few very sparse random
vectors with Bernoulli-generated nonzero entries. These sparse random
projections have been applied in many large-scale applications such
as compressive sensing and object tracking \cite{FastAndEfficientCS,FastCompressiveTracking} and to efficient learning of principal components in the large-scale data setting \cite{Farhad_ICML}. 
 To further
improve efficiency of our approach, we 
show how to
share the same random
matrix across blocks of data samples.

\section{Prior Work on Compressive Dictionary Learning \label{sec:Prior-Work}}

Several attempts have been made to address the problem of dictionary learning from compressive measurements.  In three roughly contemporary papers \cite{BlindCSTech,StuderDict}, and our work \cite{CKSVD_ICASSP}, three similar algorithms were presented to learn a dictionary based on compressive measurements.  Each was inspired by the well-known K-SVD algorithm and closely followed its structure, except in that each aimed to minimize the representation error of the compressive measurements instead of that of the original signals.  The exact steps of each algorithm have minor differences, but take a similar overall form. 

However, none of these works explicitly aimed at \emph{designing} the compressive measurements (sketches) to promote the computational efficiency of the resulting compressive K-SVD, so that it would be maximally practical for dictionary learning on large-scale data.   Moreover, none of these works gave theoretical performance analysis for such computationally-efficient sketches.  

In this paper,
we extend the previous line of work on compressive dictionary learning 
by analyzing the scheme under assumptions that make it memory and computation efficient.
The key to the efficiency of the new scheme is in considering a wider and more general class of random projection matrices for the sketches, including some very sparse ones.  We further introduce an initial analysis of the theoretical performance of compressive dictionary learning under these more general random projections.  

In this section, we review the general dictionary learning problem and the 
compressive K-SVD (CK-SVD) algorithm that was 
introduced in \cite{CKSVD_ICASSP} for the case of random Gaussian
matrices.  (We note that the approaches of \cite{StuderDict} and \cite{BlindCSTech} are similar.) Given a set of $n$ training signals $\mathbf{X}=[\mathbf{x}_{1},\ldots,\mathbf{x}_{n}]$
in $\mathbb{R}^{p}$, the dictionary learning problem is to find 
a dictionary $\mathbf{D}\in\mathbb{R}^{p\times K}$
that leads to the best representation under a strict sparsity constraint
for each member in the set, i.e., minimizing 
\begin{equation}
\min_{\D\in\RR^{p\times K},\C\in\RR^{K\times n}} \sum_{i=1}^n \left\Vert \x_i-\D\c_i\right\Vert_{2}^{2}\; s.t.\;\forall i,\;\left\Vert \c_{i}\right\Vert _{0}\leq T\label{eq:Obj_Func_Dic_Orig}
\end{equation}
where $\mathbf{C}=[\mathbf{c}_{1},\ldots,\mathbf{c}_{n}]$ is the
coefficient matrix and the $\ell_{0}$ pseudo-norm $\left\Vert \mathbf{c}_{i}\right\Vert _{0}$
counts the number of nonzero entries of the coefficient vector $\mathbf{c}_{i}\in\mathbb{R}^{K}$.
Moreover, the columns of the dictionary $\mathbf{D}=[\mathbf{d}_{1},\ldots,\mathbf{d}_{K}]$
are typically assumed to have unit $\ell_{2}$-norm. Problem \eqref{eq:Obj_Func_Dic_Orig} is generally intractable so we look for approximate solutions (e.g., via K-SVD~\cite{KSVD}).

We then consider compressed measurements (sketches), where each measurement is obtained by taking inner products of the data sample $\mathbf{x}_{i}\in\RR^p$ with
the columns of a matrix $\mathbf{R}_{i}$, i.e.,~$\mathbf{y}_{i}=\mathbf{R}_{i}^{T}\mathbf{x}_{i}$
with $\left\{ \mathbf{R}_{i}\right\} _{i=1}^{n}\in\mathbb{R}^{p\times m}$,
$m<p$, and $\mathbf{Y}=[\mathbf{y}_{1},\ldots,\mathbf{y}_{n}]\in\mathbb{R}^{m\times n}$.
In \cite{CKSVD_ICASSP}, the entries of $\R_i$ are i.i.d.~from a zero-mean Gaussian distribution,
which is an assumption we drop in the current paper.

Given access only to the compressed measurements $\y_i$ and not $\x_i$, we attempt to solve
the following compressive dictionary learning problem:
\begin{equation}
\min_{\substack{\mathbf{D}\in\mathbb{R}^{p\times K} \\\mathbf{C}\in\mathbb{R}^{K\times n}}}\;\sum_{i=1}^{n}\left\Vert \mathbf{y}_{i}-\mathbf{R}_{i}^{T}\mathbf{D}\mathbf{c}_{i}\right\Vert _{2}^{2}\; s.t.\;
\begin{cases}
\forall i,\;\left\Vert \mathbf{c}_{i}\right\Vert _{0}\leq T \\
\forall k,\;\|\mathbf{d}_k \|_2 = 1
\end{cases}
\label{eq:Obj_Func_Dic_Compressed}
\end{equation}

In the CK-SVD algorithm, the objective function in (\ref{eq:Obj_Func_Dic_Compressed})
is minimized in a simple iterative approach that alternates between
sparse coding and dictionary update steps.

\subsection{Sparse Coding}

In the sparse coding step, the penalty term in (\ref{eq:Obj_Func_Dic_Compressed})
is minimized with respect to a fixed $\mathbf{D}$ to find the coefficient
matrix $\mathbf{C}$ under the strict sparsity constraint. This can
be written as 

\begin{equation}
\min_{\mathbf{C}\in\mathbb{R}^{K\times n}}\sum_{i=1}^{n}\left\Vert \mathbf{y}_{i}-\Psi_{i}\mathbf{c}_{i}\right\Vert _{2}^{2}\; s.t.\;\forall i,\;\left\Vert \mathbf{c}_{i}\right\Vert _{0}\leq T\label{eq:SparseCode}
\end{equation}
where $\Psi_{i}=\mathbf{R}_{i}^{T}\mathbf{D}\in\mathbb{R}^{m\times K}$
is a fixed equivalent dictionary for representation of $\mathbf{y}_{i}$.
This optimization problem can be considered as $n$ distinct optimization
problems for each compressive measurement. We can then use a variety of algorithms, such as OMP,
 to find the approximate
solution $\mathbf{c}_{i}$~\cite{ComputationalTropp}.

\subsection{Dictionary Update}

The approach is to update the $k^\text{th}$ dictionary atom $\mathbf{d}_{k}$
and its corresponding coefficients 
while holding  $\mathbf{d}_{j}$ fixed for $j\neq k$,
and then repeat for $k+1$, until $k=K$.
The penalty term in (\ref{eq:Obj_Func_Dic_Compressed}) can be written
as 
\begin{eqnarray}
\hspace{-0.2in} & \hspace{-0.1in} & \hspace{-0.1in}\sum_{i=1}^{n}\Big\Vert \mathbf{y}_{i}-\mathbf{R}_{i}^{T}\sum_{j=1}^{K}c_{i,j}\mathbf{d}_{j}\Big\Vert _{2}^{2}\negthinspace\nonumber \\
\hspace{-0.2in} & \hspace{-0.1in} & \hspace{-0.1in}=\sum_{i=1}^{n}\Big\Vert \Big(\mathbf{y}_{i}-\mathbf{R}_{i}^{T}\sum_{j\neq k}c_{i,j}\mathbf{d}_{j}\Big)-c_{i,k}\mathbf{R}_{i}^{T}\mathbf{d}_{k}\Big\Vert _{2}^{2}\nonumber \\
\hspace{-0.2in} & \hspace{-0.1in} & \hspace{-0.1in}=\sum_{i\in\mathcal{I}_{k}}\Big\Vert \mathbf{e}_{i,k}-c_{i,k}\mathbf{R}_{i}^{T}\mathbf{d}_{k}\Big\Vert _{2}^{2}+\sum_{i\notin\mathcal{I}_{k}}\Big\Vert \mathbf{e}_{i,k}\Big\Vert _{2}^{2}\label{eq:DicUpCKSVD}
\end{eqnarray}
where $c_{i,k}$ is the $k^{th}$ element of $\mathbf{c}_{i}\in\mathbb{R}^{K}$,
$\mathcal{I}_{k}$ is a set of indices of compressive measurements
for which $c_{i,k}\neq 0$, and $\mathbf{e}_{i,k}=\mathbf{y}_{i}-\mathbf{R}_{i}^{T}\sum_{j\neq k}c_{i,j}\mathbf{d}_{j}\in\mathbb{R}^{m}$
is the representation error for $\mathbf{y}_{i}$ when the $k^{th}$
dictionary atom is removed. The penalty term in (\ref{eq:DicUpCKSVD})
is a quadratic function of $\mathbf{d}_{k}$ and the minimizer is
obtained by setting the derivative with respect to $\mathbf{d}_{k}$
equal to zero. Hence, 

\begin{equation}
\mathbf{G}_{k}\mathbf{d}_{k}=\mathbf{b}_{k}\label{eq:Update_Dict_CKSVD}
\end{equation}
where $\mathbf{G}_{k}=\sum_{i\in\mathcal{I}_{k}}c_{i,k}^{2}\mathbf{R}_{i}\mathbf{R}_{i}^{T}$ and $\mathbf{b}_{k}=\sum_{i\in\mathcal{I}_{k}}c_{i,k}\mathbf{R}_{i}\mathbf{e}_{i,k}$.
Therefore, we get the closed-form solution $\mathbf{d}_{k}=\mathbf{G}_{k}^{+}\mathbf{b}_{k}$, where $\mathbf{G}_{k}^{+}$ denotes the Moore-Penrose pseudo-inverse of $\mathbf{G}_{k}$.
 Once given the new $\mathbf{d}_{k}$ (normalized to have unit $\ell_{2}$-norm), the optimal $c_{i,k}$ for each $i\in\mathcal{I}_{k}$
is given by least squares as $c_{i,k}=\frac{\langle\mathbf{e}_{i,k},\mathbf{R}_{i}^{T}\mathbf{d}_{k}\rangle}{\left\Vert \mathbf{R}_{i}^{T}\mathbf{d}_{k}\right\Vert _{2}^{2}}$.
By design, the support of the coefficient matrix $\mathbf{C}$
is preserved, just as in the K-SVD algorithm.

\section{Initial Theoretical Analysis: K-Means Case\label{sec:Theoretical-Analysis}}

In this section, we provide an initial theoretical analysis of the performance of the CK-SVD algorithm by restricting our attention to a special case of dictionary learning: K-means clustering. In this special case, we can provide theoretical guarantees on the performance of CK-SVD at every step, in relation to the steps of K-means. Moreover, these guarantees will hold for a very general class of projection matrices including very sparse random projections.

We consider a statistical framework to establish the connection between
CK-SVD and K-means. K-means clustering can be viewed as a special case of dictionary learning in which each data sample is allowed
to use one dictionary atom (cluster center), i.e.~$T=1$, and the
corresponding coefficient is set to be $1$. Therefore, we consider
the following generative model 

\begin{equation}
\mathbf{x}_{i}=\overline{\mathbf{d}_{k}}+\epsilon_{i},\; i\in\mathcal{I}_{k}\label{eq:prob_model_K_means-1}
\end{equation}
where $\overline{\mathbf{d}_{k}}$ is the center of the $k^{th}$
cluster, and $\{\epsilon_{i}\}_{i=1}^{n}\in\mathbb{R}^{p}$ represent
residuals in signal approximation and they are drawn i.i.d.~from
$\mathcal{N}(\mathbf{0},\frac{\sigma^{2}}{p}\mathbf{I}_{p\times p})$,
so that the approximation error is $\mathbb{E}[\left\Vert \epsilon\right\Vert _{2}^{2}]=\sigma^{2}$. 
The set $\{\I_k\}_{k=1}^K$ is an arbitrary partition of $(1,2,\ldots, n)$,
with the condition that $|\I_k|\rightarrow \infty$ as $n\rightarrow \infty$.
The random matrices $\R_i$ are assumed to satisfy the following:
\begin{assumption}
Each entry of the random matrices $\{\mathbf{R}_{i}\}_{i=1}^{n}\in\mathbb{R}^{p\times m}$
is drawn i.i.d.~from a general class of zero-mean distributions with
finite first four moments $\{\mu_{k}\}_{k=1}^{4}$.
\label{assumption:1}
\end{assumption}
 We will see that the distribution's kurtosis is a key factor in our results. The kurtosis,
defined as $\kappa\triangleq\frac{\mu_{4}}{\mu_{2}^{2}}-3$, is a
measure of peakedness and heaviness of tail for a distribution.

We now show how, in this special case of K-means, CK-SVD would update the cluster centers.  As mentioned before, in this case, we should set $T=1$ and the correponding coefficients are set to be $1$. This
means that for all $i\in\mathcal{I}_{k}$, we have $c_{i,k}=1$, and
$c_{i,j}=0$ for $j\neq k$, and it leads to $\mathbf{e}_{i,k}=\mathbf{y}_{i},\forall i \in \mathcal{I}_k$.
Then, the update formula for the $k^{th}$ dictionary atom of CK-SVD
given in (\ref{eq:Update_Dict_CKSVD}) reduces to 

\begin{equation}
\Big(\sum_{i\in\mathcal{I}_{k}}\mathbf{R}_{i}\mathbf{R}_{i}^{T}\Big)\mathbf{d}_{k}=\sum_{i\in\mathcal{I}_{k}}\mathbf{R}_{i}\mathbf{y}_{i}.\label{eq:K-means-CK-SVD-1}
\end{equation}
Hence, similar to K-means, the process of updating $K$ dictionary
atoms becomes independent of each other. We can rewrite (\ref{eq:K-means-CK-SVD-1}) as $\mathbf{H}_{k}\mathbf{d}_{k}=\mathbf{f}_{k}$, where 
\begin{equation}
\mathbf{H}_{k}\triangleq\frac{1}{m\mu_{2}}\frac{1}{\left|\mathcal{I}_{k}\right|}\sum_{i\in\mathcal{I}_{k}}\mathbf{R}_{i}\mathbf{R}_{i}^{T},\;\mathbf{f}_{k}\triangleq\frac{1}{m\mu_{2}}\frac{1}{\left|\mathcal{I}_{k}\right|}\sum_{i\in\mathcal{I}_{k}}\mathbf{R}_{i}\mathbf{y}_{i}.\label{eq:H_K_F_K}
\end{equation}
In \cite{Farhad_ICML}, it is shown that $\mathbb{E}[\mathbf{R}_{i}\mathbf{R}_{i}^{T}]=m\mu_{2}\mathbf{I}_{p\times p}$.
Thus, we see

\begin{eqnarray}
\hspace{-0.1in}\hspace{-0.1in}\mathbb{E}\left[\mathbf{R}_{i}\mathbf{y}_{i}\right] & \hspace{-0.1in}=\hspace{-0.1in} & \mathbb{E}\left[\mathbf{R}_{i}\mathbf{R}_{i}^{T}\mathbf{x}_{i}\right]\nonumber \\
\hspace{-0.1in}\hspace{-0.1in} & \hspace{-0.1in}=\hspace{-0.1in} & \mathbb{E}\left[\mathbf{R}_{i}\mathbf{R}_{i}^{T}\overline{\mathbf{d}_{k}}\right]+\mathbb{E}\left[\mathbf{R}_{i}\mathbf{R}_{i}^{T}\epsilon_{i}\right]\nonumber \\
\hspace{-0.1in}\hspace{-0.1in} & \hspace{-0.1in}=\hspace{-0.1in} & \mathbb{E}\left[\mathbf{R}_{i}\mathbf{R}_{i}^{T}\right]\overline{\mathbf{d}_{k}}+\mathbb{E}\left[\mathbf{R}_{i}\mathbf{R}_{i}^{T}\right]\mathbb{E}\left[\epsilon_{i}\right]=m\mu_{2}\overline{\mathbf{d}_{k}}.\label{eq:expectation_RY}
\end{eqnarray}
Therefore, when the number of samples is sufficiently large, using
the law of large numbers, $\mathbf{H}_{k}$ and $\mathbf{f}_{k}$
converge to $\frac{1}{m\mu_{2}}\mathbb{E}[\mathbf{R}_{i}\mathbf{R}_{i}^{T}]=\mathbf{I}_{p\times p}$
and $\frac{1}{m\mu_{2}}\mathbb{E}[\mathbf{R}_{i}\mathbf{y}_{i}]=\overline{\mathbf{d}_{k}}$.
Hence, the updated dictionary atom in our CK-SVD is the original center
of cluster, i.e.~$\mathbf{d}_{k}=\overline{\mathbf{d}_{k}}$, exactly
as in K-means. Note that in this case even one measurement per signal
$m=1$ is sufficient. 

The following theorem characterizes convergence rates for $\mathbf{H}_{k}$ and $\mathbf{f}_{k}$ based
on various parameters such as the number of samples and the choice of random matrices. 
\begin{thm}
\label{thm:Theorem1}
Assume Assumption~\ref{assumption:1}.
Then, $\mathbf{H}_{k}$ defined in (\ref{eq:H_K_F_K}) converges to
the identity matrix $\mathbf{I}_{p\times p}$ and for any $\eta>0$,
we have 

\begin{equation}
\mathbb{P}\left(\frac{\left\Vert \mathbf{H}_{k}-\mathbf{I}_{p\times p}\right\Vert _{F}}{\left\Vert \mathbf{I}_{p\times p}\right\Vert _{F}}\leq\eta\right)\geq1-P_{0}
\end{equation}
where 

\begin{equation}
P_{0}=\frac{1}{m\left|\mathcal{I}_{k}\right|\eta^{2}}\left(\kappa+1+p\right).
\end{equation}
Also, consider the probabilistic model given in (\ref{eq:prob_model_K_means-1})
and compressive measurements $\mathbf{y}_{i}=\mathbf{R}_{i}^{T}\mathbf{x}_{i}$.
Then, $\mathbf{f}_{k}$ defined in (\ref{eq:H_K_F_K}) converges to
the center of original data and for any $\eta>0$, we have 

\begin{equation}
\mathbb{P}\left(\frac{\left\Vert \mathbf{f}_{k}-\overline{\mathbf{d}_{k}}\right\Vert _{2}}{\left\Vert \overline{\mathbf{d}_{k}}\right\Vert _{2}}\leq\eta\right)\geq1-P_{1}
\end{equation}
where

\begin{equation}
P_{1}=P_{0}+\frac{1}{\text{SNR}}\Big(P_{0}+\frac{1}{\left|\mathcal{I}_{k}\right|\eta^{2}}\Big)
\end{equation}
and the signal-to-noise ratio is defined as $\text{SNR}\triangleq\frac{\left\Vert \overline{\mathbf{d}_{k}}\right\Vert _{2}^{2}}{\sigma^{2}}$.
 
\end{thm}
We see that for a fixed error bound $\eta$, as $\left|\mathcal{I}_{k}\right|$ increases, the error probability $P_{0}$
decreases at rate $\frac{1}{\left|\mathcal{I}_{k}\right|}$. Therefore,
for any fixed $\eta>0$, the error probability $P_{0}$ goes to zero
as $\left|\mathcal{I}_{k}\right|\rightarrow\infty$. Note
that the shape of distribution, specified by the kurtosis, is an important factor. For random matrices
with heavy-tailed entries, the error probability $P_{0}$ increases.
However, $P_{0}$ gives us an explicit tradeoff between $\left|\mathcal{I}_{k}\right|$, the measurement ratio, and anisotropy in the distribution.
For example, the increase in kurtosis can be compensated by increasing
$\left|\mathcal{I}_{k}\right|$. The convergence rate analysis for $\mathbf{f}_{k}$
follows the same path. We further note that $P_{1}$ is a decreasing
function of the signal-to-noise ratio and as $\text{SNR}$ increases,
$P_{1}$ gets closer to $P_{0}$, where for the case that $\text{SNR}\rightarrow\infty$,
then $P_{1}\thickapprox P_{0}$. 

\begin{figure}

\centering{}\includegraphics[scale=0.32]{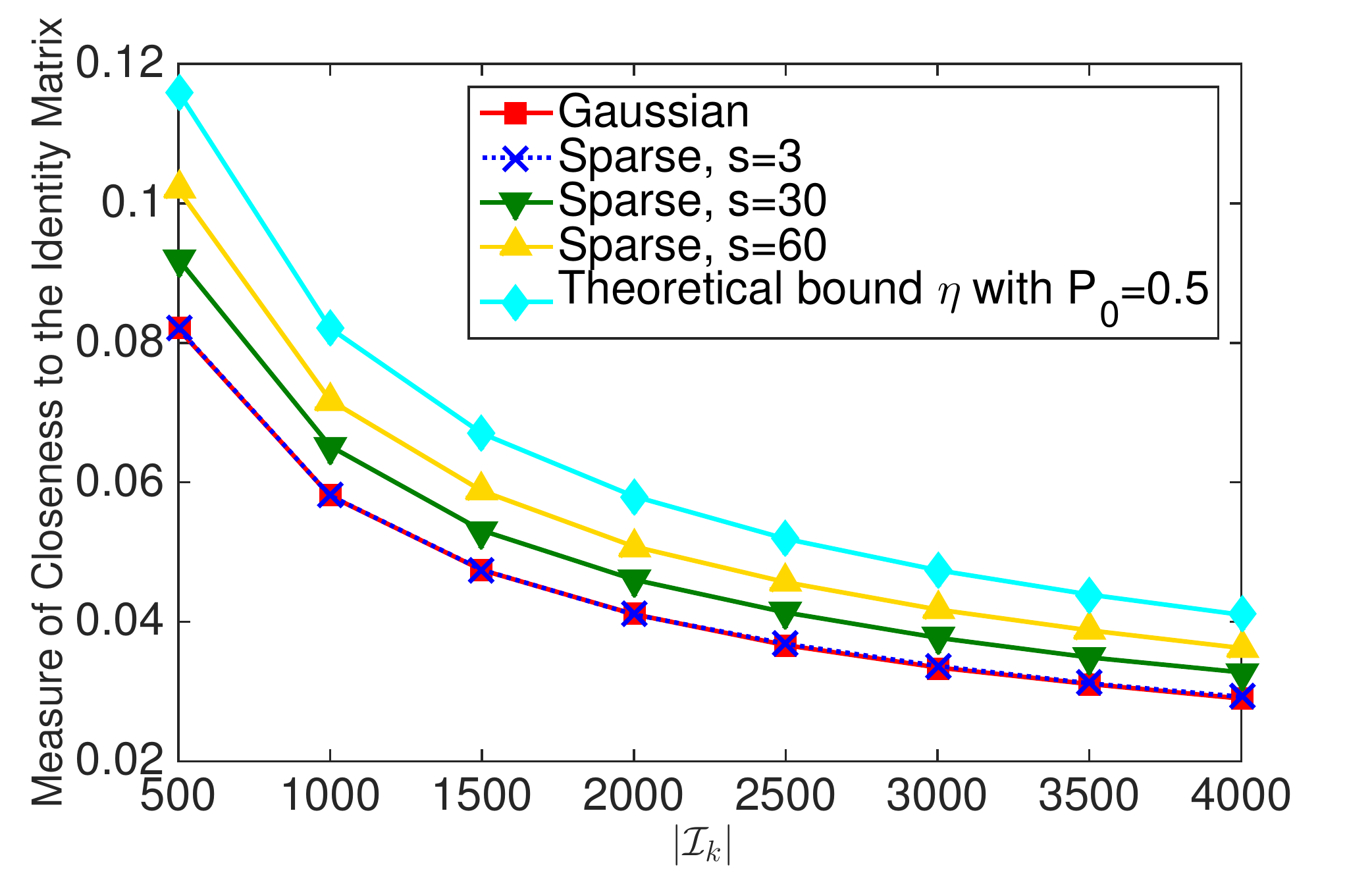}\caption{\label{fig:Closeness-Hk}Closeness of $\mathbf{H}_{k}$ to $\mathbf{I}_{p\times p}$
defined as $\left\Vert \mathbf{H}_{k}-\mathbf{I}_{p\times p}\right\Vert _{F}/\left\Vert \mathbf{I}_{p\times p}\right\Vert _{F}$.
$\{\mathbf{R}_{i}\}_{i=1}^{n}\in\mathbb{R}^{p\times m}$,
$p=100$ and $\nicefrac{m}{p}=0.3$, are generated with i.i.d.~entries
both for Gaussian and the sparse-Bernoulli distribution. We see that as $\left|\mathcal{I}_{k}\right|$ increases, $\mathbf{H}_{k}$ gets closer to $\mathbf{I}_{p\times p}$.
Also, for fixed $\left|\mathcal{I}_{k}\right|$, as the sparsity of random matrices
increases, the kurtosis $\kappa=s-3$ increases and consequently the
distance between $\mathbf{H}_{k}$ and $\mathbf{I}_{p\times p}$ increases.
For Gaussian and the sparse-Bernoulli with $s=3$,
we have $\kappa=0$. We also plot the theoretical bound $\eta$ with $P_{0}=0.5$ for the Gaussian case.}
\end{figure}

Let's consider an example to gain intuition on the choice of random
matrices. We are interested in comparing the dense
random Gaussian matrices with very sparse random matrices, where
each entry is drawn from $\{-1,0,+1\}$ with probabilities $\{\frac{1}{2s},1-\frac{1}{s},\frac{1}{2s}\}$
for $s\geq1$ (we refer to this distribution as a sparse-Bernoulli
distribution with parameter $s$). $\{\mathbf{R}_{i}\}_{i=1}^{n}\in\mathbb{R}^{p\times m}$,
$p=100$ and $\nicefrac{m}{p}=0.3$, are generated with i.i.d.~entries
both for Gaussian and the sparse-Bernoulli distribution. In Fig.~\ref{fig:Closeness-Hk}, we see
that as $\left|\mathcal{I}_{k}\right|$ increases, $\mathbf{H}_{k}$ gets
closer to the identity matrix $\mathbf{I}_{p\times p}$. Also,
for a fixed $\left|\mathcal{I}_{k}\right|$, as the sparsity of random matrices increases, the
kurtosis $\kappa=s-3$ increases. Therefore, based on Theorem \ref{thm:Theorem1},
we expect that the distance between $\mathbf{H}_{k}$ and $\mathbf{I}_{p\times p}$
increases. Note that for Gaussian and the sparse-Bernoulli
with $s=3$, we have $\kappa=0$. 

As a final note, our theoretical analysis gives us valuable insight
about the number of distinct random matrices required. Based on
Theorem \ref{thm:Theorem1}, there is an inherent tradeoff between
the accuracy and the number of distinct random
matrices used. For example, if we only use one random matrix, we are not able to
recover the true dictionary as observed in \cite{BlindCSFirst}. Also, increasing the number of distinct random matrices improves the accuracy, as mentioned in \cite{BlindCSTech}.
Hence, we can reduce the number of distinct
random matrices in large-scale problems where $n=O(p)$ with controlled loss in accuracy.

\section{Memory and Computation Efficient Dictionary Learning\label{sec:Memory-and-Computation}}
Now, we return our attention to general dictionary learning.  Inspired by the generality of the projection matrices in Theorem~\ref{thm:Theorem1}, we sketch using very sparse random matrices, and furthermore reduce the number of distinct random matrices to increase the efficiency of our approach. 

Assume that the original data samples are divided into $L$ blocks
$\mathbf{X}=[\mathbf{X}^{(1)},\ldots,\mathbf{X}^{(L)}]$, where $\mathbf{X}^{(l)}$
represents the $l^{th}$ block. Let $\mathbf{R}_{l}\in\mathbb{R}^{p\times m}$,
$m<p$, represent the random matrix used for the $l^{th}$ block.
Then, we have 

\begin{equation}
\mathbf{Y}^{(l)}=\mathbf{R}_{l}^{T}\mathbf{X}^{(l)},\;1\leq l\leq L
\end{equation}
where $\mathbf{Y}^{(l)}$ is the sketch of $\mathbf{X}^{(l)}$. Each
entry of $\{\mathbf{R}_{l}\}_{l=1}^{L}$ is distributed on $\{-1,0,+1\}$
with probabilities $\{\frac{1}{2s},1-\frac{1}{s},\frac{1}{2s}\}$.
Here, the parameter $s$ controls the sparsity of
random matrices such that each column of $\{\mathbf{R}_{l}\}_{l=1}^{L}$
has $\frac{p}{s}$ nonzero entries, on average. We are specifically
interested in choosing $m$ and $s$ such that the \textit{compression factor}
$\gamma\triangleq\frac{m}{s}<1$. Thus, the cost to acquire each compressive
measurement is $O(\gamma p)$, $\gamma<1$, vs.~the cost for collecting
every data entry $O(p)$. 

Similarly, we aim to minimize the representation error as 
\vspace{-0.05in}
\begin{equation}
\min_{\mathbf{D}\in\mathbb{R}^{p\times K},\mathbf{C}\in\mathbb{R}^{K\times n}}\sum_{l=1}^{L}\left\Vert \mathbf{Y}^{(l)}-\mathbf{R}_{l}^{T}\mathbf{D}\mathbf{C}^{(l)}\right\Vert _{F}^{2}\;\hspace{-1mm} s.t.\hspace{-1mm}\;\forall i,\;\left\Vert \mathbf{c}_{i}^{(l)}\right\Vert _{0}\hspace{-2mm}\leq T\label{eq:Obj_Func_Dic_Eff}
\end{equation}
where $\mathbf{c}_{i}^{(l)}$ represents the $i^{th}$ sample in the
$l^{th}$ block of the coefficient matrix $\mathbf{C}^{(l)}$. As before, the
penalty term in (\ref{eq:Obj_Func_Dic_Eff}) is minimized in a simple
iterative approach involving two steps.  
The first step, sparse coding, is the same as the CK-SVD algorithm previously described, except we can take efficient of the block structure and use
Batch-OMP \cite{EfficientKSVD} in each block which is significantly faster
than OMP for each $\mathbf{c}_{i}$ separately.

\subsection{Dictionary Update }

 The goal is to update the $k^{th}$ dictionary atom $\mathbf{d}_{k}$ 
for $k=1,\ldots,K$,
while assuming that $\mathbf{d}_{j}$, $j\neq k$, is fixed.
 The
penalty term in (\ref{eq:Obj_Func_Dic_Eff}) can be written as 
\vspace{-0.05in}
\begin{eqnarray}
 &  & \hspace{-6mm}\sum_{l=1}^{L}\Big\Vert \mathbf{Y}^{(l)}-\mathbf{R}_{l}^{T}\mathbf{D}\mathbf{C}^{(l)}\Big\Vert _{F}^{2}\hspace{-1mm}=\hspace{-1mm}\sum_{l=1}^{L}\sum_{i=1}^{n_{l}}\Big\Vert \mathbf{y}_{i}^{(l)}-\mathbf{R}_{l}^{T}\sum_{j=1}^{K}c_{i,j}^{(l)}\mathbf{d}_{j}\Big\Vert _{2}^{2}\nonumber \\
 &  & \hspace{-6mm}=\hspace{-1mm}\sum_{l=1}^{L}\sum_{i=1}^{n_{l}}\Big\Vert \Big(\mathbf{y}_{i}^{(l)}-\mathbf{R}_{l}^{T}\sum_{j\neq k}c_{i,j}^{(l)}\mathbf{d}_{j}\Big)-c_{i,k}^{(l)}\mathbf{R}_{l}^{T}\mathbf{d}_{k}\Big\Vert _{2}^{2}\nonumber \\
 &  & \hspace{-6mm}=\hspace{-1mm}\sum_{l=1}^{L}\sum_{i=1}^{n_{l}}\Big\Vert \mathbf{e}_{i,k}^{(l)}-c_{i,k}^{(l)}\mathbf{R}_{l}^{T}\mathbf{d}_{k}\Big\Vert _{2}^{2}\label{eq:CKSVD_Dic_Update}
\end{eqnarray}
where $c_{i,k}^{(l)}$ is the $k^{th}$ element of 
$\mathbf{c}_{i}^{(l)}\in\mathbb{R}^{K}$, and $\mathbf{e}_{i,k}^{(l)}\in\mathbb{R}^{m}$
is the representation error for the compressive measurement $\mathbf{y}_{i}^{(l)}$
when the $k^{th}$ dictionary atom is removed. The objective function
in (\ref{eq:CKSVD_Dic_Update}) is a quadratic function of $\mathbf{d}_{k}$
and the minimizer is obtained by setting the derivative of the objective
function with respect to $\mathbf{d}_{k}$ equal to zero. First, let
us define $\mathcal{I}_{k}^{(l)}$ as a set of indices of compressive
measurements in the $l^{th}$ block using $\mathbf{d}_{k}$. Therefore,
we get the following expression

\begin{equation}
\mathbf{G}_{k}\mathbf{d}_{k}=\mathbf{b}_{k},\;
\mathbf{G}_{k}\triangleq\sum_{l=1}^{L}s_{k}^{(l)}\mathbf{R}_{l}\mathbf{R}_{l}^{T},\; 
\mathbf{b}_{k}\triangleq\sum_{l=1}^{L}\sum_{i\in\mathcal{I}_{k}^{(l)}}c_{i,k}^{(l)}\mathbf{R}_{l}\mathbf{e}_{i,k}^{(l)}
\end{equation}
where
 $s_{k}^{(l)}$ is defined as the sum of squares of all the coefficients
related to the $k^{th}$ dictionary atom in the $l^{th}$ block, i.e.~$s_{k}^{(l)}\triangleq\sum_{i\in\mathcal{I}_{k}^{(l)}}(c_{i,k}^{(l)})^{2}$.

Note that $\mathbf{G}_k$ can be computed efficiently: concatenate
$\{\mathbf{R}_{l}\}_{l=1}^{L}$ 
 in a matrix  $\mathbf{R}\triangleq\left[\mathbf{R}_{1},\mathbf{R}_{2},\ldots,\mathbf{R}_{L}\right]\in\mathbb{R}^{p\times(mL)}$,
 and define
  the diagonal matrix $\mathbf{S}_{k}$ as 

\begin{equation}
\mathbf{S}_{k}\triangleq\text{diag}\Big(\Big[\underbrace{s_{k}^{(1)},\ldots,s_{k}^{(1)}}_{\text{repeated \ensuremath{m} times}},\ldots,\underbrace{s_{k}^{(L)},\ldots,s_{k}^{(L)}}_{\text{repeated \ensuremath{m} times}}\Big]\Big)
\end{equation}
where $\text{diag}(\mathbf{z})$ represents a square diagonal matrix
with the elements of vector $\mathbf{z}$ on the main diagonal. Then, we have $\mathbf{G}_{k}=\mathbf{R}\mathbf{S}_{k}\mathbf{R}^{T}$.

Given the updated $\mathbf{d}_{k}$, the optimal $c_{i,k}^{(l)}$,
for all $i\in\mathcal{I}_{k}^{(l)}$, is given by least squares as
%
$
c_{i,k}^{(l)}=\frac{\langle\mathbf{e}_{i,k}^{(l)},\mathbf{R}_{l}^{T}\mathbf{d}_{k}\rangle}{\left\Vert \mathbf{R}_{l}^{T}\mathbf{d}_{k}\right\Vert _{2}^{2}},\;\forall i\in\mathcal{I}_{k}^{(l)}.
$

\section{Experimental Results\label{sec:Experimental-Results}}

\begin{figure}
\subfloat[]{
\includegraphics[width=.24\textwidth]{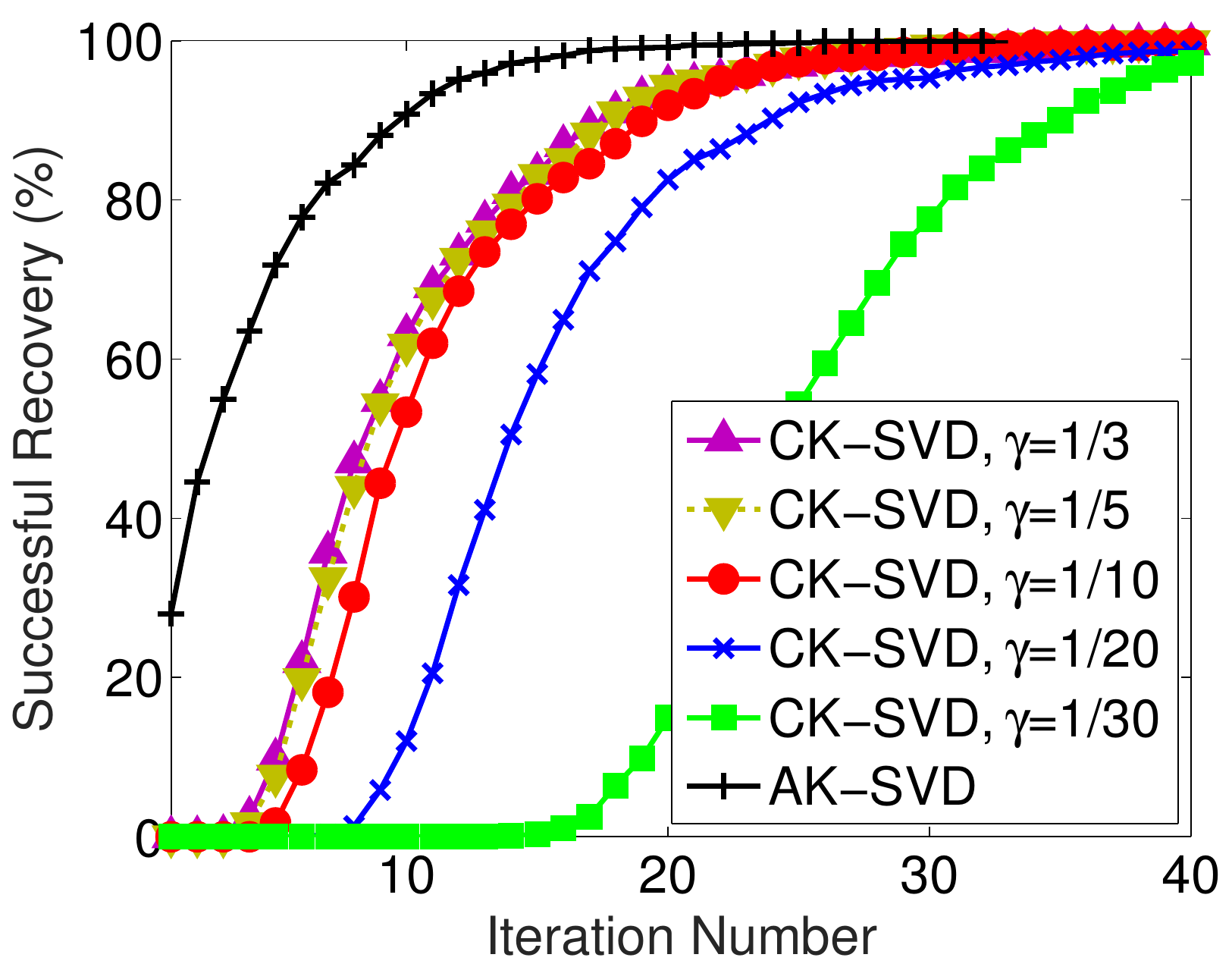}
}
\subfloat[]{
\includegraphics[width=.24\textwidth]{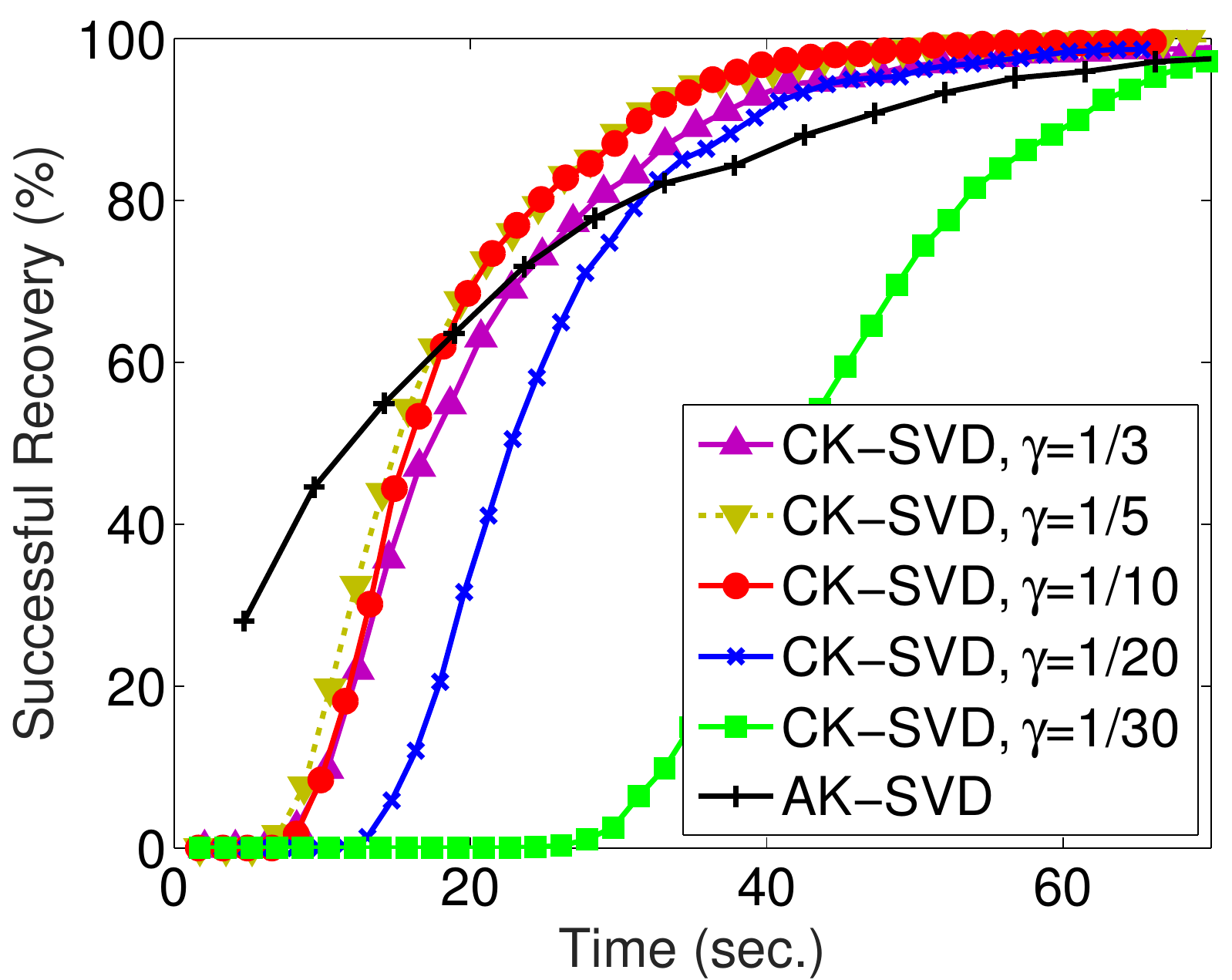}
}

\caption{\label{fig:Successful_Recovery}Results for synthetic data. Plot of
successful recovery vs.~(a) iteration number, and (b) time. Our method
CK-SVD for varying compression factor $\gamma$ is compared with
AK-SVD. We observe that our method is \emph{both} memory/computation efficient
and accurate for $\gamma=\frac{1}{5}$ and $\gamma=\frac{1}{10}$.
 }

\end{figure}

We examine the performance of our dictionary learning algorithm on
a synthetic dataset. Our proposed method is compared with the fast
and efficient implementation of K-SVD known as Approximate K-SVD (AK-SVD)
\cite{EfficientKSVD} that requires access to the entire data. We
generate $K=15$ dictionary atoms in $\mathbb{R}^{p}$, $p=1000$,
drawn from the uniform distribution and normalized to have unit norm.
A set of data samples $\{\mathbf{x}_{i}\}_{i=1}^{50,000}\in\mathbb{R}^{p}$
is generated where each sample is a linear combination of three distinct
atoms, i.e.~$T=3$, and the corresponding coefficients are chosen
i.i.d.~from the Gaussian distribution $\mathcal{N}\left(0,100\right)$.
Then, each data is corrupted by Gaussian noise drawn from $\mathcal{N}(0,0.04\mathbf{I}_{p\times p})$.

CK-SVD is applied on the set of compressive measurements obtained
by very sparse random matrices for various values of the compression
factor $\gamma=\frac{1}{3},\frac{1}{5},\frac{1}{10},\frac{1}{20},\frac{1}{30}$.
We set 
the number of blocks $L=250$ and $\nicefrac{m}{p}=0.1$.
Performance is evaluated by the magnitude of the inner product
between learned and true atoms. 
 A value greater than $0.95$
is counted as a successful recovery. Fig.~\ref{fig:Successful_Recovery}
shows the results of CK-SVD averaged over $50$ independent trials.
In practice, when $T$ is small, the updates for $\mathbf{d}_k$ are nearly decoupled, and we may delay updating $c_{i,k}^{(l)}$ until after all $K$  updates of $\mathbf{d}_k$. For $T=3$, the accuracy results are indistinguishable.

In Fig.~\ref{fig:Successful_Recovery}, 
we see that our method is able to eventually reach high accuracy even for $\gamma=\frac{1}{30}$,
achieving substantial savings in memory/data access. However, there
is a tradeoff between memory and computation savings vs.~accuracy.
Our method is efficient in memory/computation and, at the same
time, accurate for $\gamma=\frac{1}{5}$ and $\gamma=\frac{1}{10}$,
where it outperforms AK-SVD if the time of each iteration is factored in. 
We compare with AK-SVD to give an idea of our efficiency, but note
that AK-SVD and our CK-SVD are not completely comparable. 
In our example, both methods reach $100\%$ accuracy eventually but in general they may give different levels of accuracy. The main advantage of CK-SVD appears as the dimensions grow, since then memory/data access is a dominant issue.

\section*{Acknowledgment}

This material is based upon work supported by the National Science
Foundation under Grant CCF-1117775.
This work utilized the Janus supercomputer, which is supported by the National Science Foundation (award number CNS-0821794) and the University of Colorado Boulder. The Janus supercomputer is a joint effort of the University of Colorado Boulder, the University of Colorado Denver and the National Center for Atmospheric Research.

\bibliographystyle{IEEEtran}
\bibliography{CKSVD}

\end{document}